\documentclass[letterpaper]{llncs}
\usepackage{times}
\usepackage{helvet}
\usepackage{courier}
\usepackage{epic}
\usepackage{graphicx}
\usepackage{latexsym}
\usepackage{verbatim}
\usepackage{xspace}
\begin{document}

\newlength{\halftextwidth}
\setlength{\halftextwidth}{0.47\textwidth}
\def\halffigsize{2.2in}
\def\thirdfigsize{1.5in}
\def\negvspace{0in}
\def\posvspace{0em}






\newcommand{\set}{\mathcal}
\newcommand{\myset}[1]{\ensuremath{\mathcal #1}}

\renewcommand{\theenumii}{\alph{enumii}}
\renewcommand{\theenumiii}{\roman{enumiii}}
\newcommand{\figref}[1]{Figure \ref{#1}}
\newcommand{\tref}[1]{Table \ref{#1}}
\newcommand{\myOmit}[1]{}
\newcommand{\And}{\wedge}
\newcommand{\myldots}{.}

\newcommand{\nina}[1]{{#1}}

\newtheorem{mydefinition}{Definition}
\newtheorem{mytheorem}{Theorem}
\newtheorem{mycor}{Corollary}
\newenvironment{myexample}{{\bf Running example:} \it}{\rm}
\newtheorem{mytheorem1}{Theorem}
\newcommand{\myproof}{\noindent {\bf Proof:\ \ }}
\newcommand{\myqed}{\mbox{$\Box$}}

\newcommand{\mymod}{\mbox{\rm mod}}
\newcommand{\range}{\mbox{\sc Range}}
\newcommand{\roots}{\mbox{\sc Roots}}
\newcommand{\myiff}{\mbox{\rm iff}}
\newcommand{\alldifferent}{\mbox{\sc AllDifferent}}
\newcommand{\alldiff}{\mbox{\sc AllDifferent}}
\newcommand{\interdistance}{\mbox{\sc InterDistance}}
\newcommand{\permutation}{\mbox{\sc Permutation}}
\newcommand{\disjoint}{\mbox{\sc Disjoint}}
\newcommand{\cardpath}{\mbox{\sc CardPath}}
\newcommand{\CARDPATH}{\mbox{\sc CardPath}}
\newcommand{\knapsack}{\mbox{\sc Knapsack}}
\newcommand{\common}{\mbox{\sc Common}}
\newcommand{\uses}{\mbox{\sc Uses}}
\newcommand{\lex}{\mbox{\sc Lex}}
\newcommand{\LEX}{\mbox{\sc Lex}}
\newcommand{\SnakeLex}{\mbox{\sc SnakeLex}}
\newcommand{\usedby}{\mbox{\sc UsedBy}}
\newcommand{\nvalue}{\mbox{\sc NValue}}
\newcommand{\slide}{\mbox{\sc Slide}}
\newcommand{\SLIDE}{\mbox{\sc Slide}}
\newcommand{\circularslide}{\mbox{\sc Slide}_{\rm O}}
\newcommand{\among}{\mbox{\sc Among}}
\newcommand{\mysum}{\mbox{\sc Sum}}
\newcommand{\amongseq}{\mbox{\sc AmongSeq}}
\newcommand{\atmost}{\mbox{\sc AtMost}}
\newcommand{\atleast}{\mbox{\sc AtLeast}}
\newcommand{\element}{\mbox{\sc Element}}
\newcommand{\gcc}{\mbox{\sc Gcc}}
\newcommand{\egcc}{\mbox{\sc EGcc}}
\newcommand{\gsc}{\mbox{\sc Gsc}}
\newcommand{\contiguity}{\mbox{\sc Contiguity}}
\newcommand{\PRECEDENCE}{\mbox{\sc Precedence}}
\newcommand{\precedence}{\mbox{\sc Precedence}}
\newcommand{\assignnvalues}{\mbox{\sc Assign\&NValues}}
\newcommand{\linksettobooleans}{\mbox{\sc LinkSet2Booleans}}
\newcommand{\domain}{\mbox{\sc Domain}}
\newcommand{\symalldiff}{\mbox{\sc SymAllDiff}}
\newcommand{\valsymbreak}{\mbox{\sc ValSymBreak}}
\newcommand{\RowColSym}{\mbox{\sc RowWiseLexLeader}}
\newcommand{\RowColSymShort}{\mbox{\sc RowWiseLex}}
\newcommand{\RowSymShort}{\mbox{\sc RowLex}}
\newcommand{\LexLeader}{\mbox{\sc LexLeader}}
\newcommand{\ColSymShort}{\mbox{\sc ColLex}}
\newcommand{\NoSymShort}{\mbox{\sc NoSB}}
\newcommand{\RowLexLeader}{\mbox{\sc RowLexLeader}}
\newcommand{\LexChain}{\mbox{\sc LexChain}}
\newcommand{\OrderRowCol}{\mbox{\sc Order1stRowCol}}

\newcommand{\slidingsum}{\mbox{\sc SlidingSum}}
\newcommand{\MaxIndex}{\mbox{\sc MaxIndex}}
\newcommand{\REGULAR}{\mbox{\sc Regular}}
\newcommand{\regular}{\mbox{\sc Regular}}
\newcommand{\Regular}{\mbox{\sc Regular}}
\newcommand{\STRETCH}{\mbox{\sc Stretch}}
\newcommand{\SLIDEOR}{\mbox{\sc SlideOr}}
\newcommand{\NAE}{\mbox{\sc NotAllEqual}}
\newcommand{\mymax}{\mbox{\rm max}}

\newcommand{\todo}[1]{{\tt (... #1 ...)}}

\newcommand{\DC}{\ensuremath{DC}\xspace}
\newcommand{\Xbf}{\mbox{{\bf X}}\xspace}
\newcommand{\LEXCHAIN}{\mbox{\sc LexChain}}
\newcommand{\DLex}{\mbox{\sc DoubleLex}\xspace}
\newcommand{\snakelex}{\mbox{\sc SnakeLex}\xspace}
\newcommand{\DLexColSum}{\mbox{\sc DoubleLexColSum}\xspace}

\title{On The Complexity and Completeness of Static Constraints for
  Breaking Row and Column Symmetry\thanks{
Supported by ANR UNLOC project, ANR 08-BLAN-0289-01
and the Australian 
Government's  Department of Broadband, Communications and the Digital Economy
and the
ARC.}}
\author{
George Katsirelos\inst{1} \and
Nina Narodytska\inst{2}
\and 
Toby Walsh\inst{2}}
\institute{CRIL-CNRS, Lens, France, email: gkatsi@gmail.com
\and NICTA and University of NSW,
Sydney, Australia, email: 
\{nina.narodytska,toby.walsh\}@nicta.com.au
}
\myOmit{

\title{The Complexity of Lexicographic Ordering Constraints for Breaking Row and Column Symmetry}
\author{Keywords:
\And
Constraint Satisfaction (General/other),
\And
Global Constraints
}
}

\maketitle
\begin{abstract}
We consider a common type of symmetry
where we have 
a matrix of decision variables
with interchangeable rows and columns. 
A simple and efficient
method to deal with such row and column symmetry
is to post symmetry breaking constraints like \DLex
and \snakelex. 
We provide a number of
positive and negative results on posting
such symmetry breaking constraints. 
On the positive side, we prove that
we can compute in polynomial time
a unique representative of an
equivalence class in a matrix model
with row and column symmetry
if the number of rows (or of columns)
is bounded and in
a number of other special cases. 
On the negative side, we show that whilst
\DLex and \snakelex are often effective in practice,
they can leave a large
number of 
symmetric
solutions in the worst case. In addition, we prove that propagating
\DLex completely is NP-hard.
Finally we consider how to break 
row, column
and value symmetry, correcting a result in the literature
about the safeness of combining different symmetry 
breaking constraints. 
We end with the first experimental study on how
much symmetry is left by \DLex 
and \snakelex on some benchmark problems. 
\end{abstract}

\myOmit{

}

\section{Introduction}

One challenge in constraint programming
is to develop effective search methods
to deal with common modelling patterns. 
One such pattern is row and column symmetry \cite{ffhkmpwcp2002}:
many problems can be modelled by a matrix 
of decision variables \cite{matrix} where the rows and columns
of the matrix are fully or partially interchangeable.
Such symmetry is a source of combinatorial complexity. 
It is therefore important to develop techniques to deal with
this type of symmetry. We study here simple constraints that can be posted
to break row and column symmetries, and analyse their effectiveness 
both theoretically and experimentally. We prove that
we can compute in polynomial time
the lexicographically smallest 
representative of an
equivalence class in a matrix model
with row and column symmetry
if the number of rows (or of columns)
is bounded and thus remove all symmetric
solutions.
We are therefore able for the first
time to see how much symmetry is left by 
these commonly used symmetry breaking constraints. 

\section{Formal background}

A constraint satisfaction problem (CSP) consists of a set of
variables, each with a domain of values, and a set of
constraints specifying allowed values for 
subsets of variables. 
When solving a CSP, we often use
propagation algorithms to prune the search
space by enforcing properties like domain 
consistency. A constraint is \emph{domain
consistent} (\emph{DC})
iff when a variable in the scope of a constraint
is assigned any value in its domain, there
exist compatible values in the domains of all the other variables 
in the scope of the constraint. 
A CSP is domain consistent iff every
constraint is domain consistent.
An important feature of many
CSPs is symmetry.
Symmetries can act on variables or values (or both). 
A \emph{variable symmetry} is a bijection $\sigma$ on
the variable indices that preserves solutions. 
That is, if $\{X_i = a_i \ | \ i \in [1,n]\}$
is a solution then $\{X_{\sigma(i)} = a_i \ | \ i \in [1,n]\}$ is also.
A \emph{value symmetry} is a bijection $\theta$ on
the values that preserves solutions. 
That is, if $\{X_i = a_i \ | \ i \in [1,n]\}$
is a solution then $\{X_{i} = \theta(a_i) \ | \ i \in [1,n]\}$ is also.
A simple but effective method to deal with 
symmetry is to add \emph{symmetry breaking constraints} 
which eliminate 
symmetric solutions. 
For example, 
Crawford {\it et al.} proposed the general lex-leader method
that posts lexicographical ordering constraints 
to eliminate all but the lexicographically
least solution in each symmetry class \cite{clgrkr96}. 
%
%
Many problems are naturally modelled 
by a matrix of decision variables with
variable symmetry in which
the rows and/or columns are
interchangeable \cite{ffhkmpwcp2002}.
We say that a CSP containing
a matrix of decision variables has row symmetry iff 
given a solution, any permutation of the rows is
also a solution.
Similarly, it has column symmetry iff 
given a solution, any permutation of the columns is
also a solution.

\begin{myexample}
The Equidistant Frequency Permutation Array (EFPA) problem
is a challenging problem in coding theory. The goal
is to find a set of $v$ code words, each of length $q\lambda$
such that each word contains $\lambda$ copies of the symbols
1 to $q$, and each pair of code words is Hamming distance
$d$ apart. For example, for $v=4$, $\lambda=2$, $q=3$, $d=4$,
one solution is:
\alph{equation}
\renewcommand{\theequation}{\alph{equation}}
\begin{equation} \label{epfa}
\begin{array}{cccccc}
0 & 2 & 1 & 2 & 0 & 1 \\
0 & 2 & 2 & 1 & 1 & 0 \\
0 & 1 & 0 & 2 & 1 & 2 \\
0 & 0 & 1 & 1 & 2 & 2  
\end{array}
\end{equation}
This problem has applications in 
communication theory,
and is 
related to other combinatorial
problems like finding 
orthogonal Latin squares.
Huczynska {\it et al.} \cite{hmmncp09} consider a 
model for this problem with a $v$ by $q\lambda$
array of variables with domains $1$ to $q$. 
This model has row and column symmetry since we can
permute the rows and columns and still have
a solution. 
\end{myexample}

\section{Breaking row and column symmetry}

To break all row symmetry we can
post lexicographical ordering constraints
on the rows. Similarly,
to break all column symmetry we can 
post lexicographical ordering constraints
on the columns. 
When we have both row and column symmetry,
we can post 
a $\DLex$ constraint that 
lexicographically orders
both the rows and columns
\cite{ffhkmpwcp2002}.
This does not eliminate
all symmetry since
it may not break symmetries which permute 
both rows and columns. Nevertheless, it is often
effective in practice. 

\begin{myexample}
Consider again solution 
\mbox{\rm (\ref{epfa})}.
If we order the rows of \mbox{\rm (\ref{epfa})} lexicographically, we 
get a solution with lexicographically
ordered rows and columns:
\renewcommand{\theequation}{\alph{equation}}
\begin{equation} \label{epfa2}
\begin{array}{cccccc}
0 & 2 & 1 & 2 & 0 & 1 \\
0 & 2 & 2 & 1 & 1 & 0 \\
0 & 1 & 0 & 2 & 1 & 2 \\
0 & 0 & 1 & 1 & 2 & 2  
\end{array}
\
\begin{array}{c}
\mbox{\rm order} \\
\ \ \ \ \Rightarrow \ \ \ \ \\
\mbox{\rm rows}
\end{array}
\ \
\begin{array}{cccccc}
0 & 0 & 1 & 1 & 2 & 2 \\
0 & 1 & 0 & 2 & 1 & 2 \\
0 & 2 & 1 & 2 & 0 & 1 \\
0 & 2 & 2 & 1 & 1 & 0 
\end{array}
\end{equation}
Similarly 
if we order the columns of \mbox{\rm (\ref{epfa})} lexicographically, we 
get a different solution in which both rows and columns
are again ordered lexicographically:
\renewcommand{\theequation}{\alph{equation}}
\begin{equation} \label{epfa3}
\begin{array}{cccccc}
0 & 2 & 1 & 2 & 0 & 1 \\
0 & 2 & 2 & 1 & 1 & 0 \\
0 & 1 & 0 & 2 & 1 & 2 \\
0 & 0 & 1 & 1 & 2 & 2  
\end{array}
\ 
\begin{array}{c}
\mbox{\rm order} \\
\ \ \ \ \Rightarrow \ \ \ \ \\
\mbox{\rm cols}
\end{array}
\ \ 
\begin{array}{cccccc}
0 & 0 & 1 & 1 & 2 & 2 \\
0 & 1 & 0 & 2 & 1 & 2 \\
0 & 1 & 2 & 0 & 2 & 1 \\
0 & 2 & 2 & 1 & 1 & 0  
\end{array}
\end{equation}

All three solutions are thus in the same row and column symmetry 
class. However, both 
\renewcommand{\theequation}{\alph{equation}}
\mbox{\rm (\ref{epfa2})} and 
\mbox{\rm (\ref{epfa3})}
satisfy the \DLex constraint.
Therefore \DLex can leave multiple
solutions in each symmetry class. 
\end{myexample}

The lex-leader
method breaks all symmetry by
ensuring that any solution is
the lexicographically
smallest in its symmetry class \cite{clgrkr96}. 
This requires linearly ordering
the matrix. Lexicographically
ordering the rows and columns is consistent with
a linearization that takes the matrix in row-wise 
order (i.e. appending rows in order). 
We therefore consider a complete symmetry breaking
constraint \RowColSym\ which ensures that the row-wise
linearization of the matrix is lexicographically 
smaller than 
all its row or column permutations,
or compositions of row and column
permutations. 

\renewcommand{\theequation}{\alph{equation}}

\begin{myexample}
Consider the symmetric solutions 
\mbox{\rm (\ref{epfa})}
to \mbox{\rm (\ref{epfa3})}. If we linearize
these solutions row-wise, 
the first two are lexicographically
larger than the third. Hence, the
first two solutions are eliminated by 
the \RowColSym\ constraint. 
\end{myexample}

\RowColSym\ breaks all row and column symmetries. 
Unfortunately, posting such a constraint
is problematic since it is NP-hard to check 
if a complete assignment satisfies \RowColSym\ 
\cite{bhhwaaai2004,complexity}. We now give our
first major result. We prove that
if we can bound the number of rows (or columns),
then there is a polynomial time method to break all
row and column symmetry. For example,
in the EFPA problem, the number of columns
might equal the fixed word size of our computer. 

\begin{mytheorem}
\label{tm:fpt}
For a $n$ by $m$ matrix, 
we can check if a complete assignment satisfies a \RowColSym\ constraint 
in $O(n! nm \log m)$ time.
\end{mytheorem}
\myproof
Consider the matrix model $X_{i,j}$. 
\nina{
We exploit the fact that with no row symmetry and
just column symmetry, 
lexicographically ordering the columns gives the lex-leader 
assignment. }
\nina{
Let $Y_{i,j}= X_{\sigma(i),j}$ be a row permutation of $X_{i,j}$.
To obtain $Z_{i,j}$,
the smallest column permutation of  $Y_{i,j}$ 
we lexicographically sort the $m$ columns of $Y_{i,j}$ 
in $O(n m \log(m))$ time. Finally, we check that 
$[X_{1,1},\ldots,X_{1,m},\ldots,X_{n,1},\ldots,X_{n,m}] 
\leq_{\rm lex} [Z_{1,1},\ldots,Z_{1,m},\ldots,Z_{n,1},\ldots,Z_{n,m}]$,
where $\leq_{\rm lex}$ is the lexicographic comparison of two
vectors.
This ensures that $X_{i,j}$ is lexicographically smaller
than or equal to any column permutation of this row permutation. 
}
If we do this for each of the 
$n!-1$ non-identity row
permutations, then $X_{i,j}$ is lexicographically smaller
than or equal to any row permutation. This
means that we have the lex-leader assignment. 
This can be done in time $O(n! nm\log m)$, which
for bounded $n$ is polynomial.
%
%
\myqed

This result easily generalizes to
when rows and columns 
are partially interchangeable.
In the experimental section, we show
that this gives an effective method to
break {\em all} row and column symmetry. 

\section{Double Lex}

When the number of both rows and columns is large,
breaking {all} row and column symmetry is
computationally challenging. In this situation,
we can post a \DLex constraint \cite{ffhkmpwcp2002}.
However, as we saw in the running example, this may not break all
symmetry. 
In fact, 
it can leave $n!$ symmetric solutions
in an $2n\times 2n$ matrix model.

\begin{mytheorem}
\label{tm:expsol}
There exists a class of $2n$ by $2n$
0/1 matrix models on which
\DLex leaves $n!$ symmetric solutions,
for all $n \geq 2$.
\end{mytheorem}
\myproof
Consider a $2n$ by $2n$ matrix model with
the constraints that the matrix contains 
$3n$ non-zero entries, and each row
and column contains between one and
two non-zero entries. This model has row and column symmetry
since row and column permutations leave
the constraints unchanged.
There exists a class of symmetric solutions
to the problem that satisfy a \DLex
constraint of the form:
$$
\begin{array}{cc}
0 & I^R \\
I^R & P \end{array}
$$
Where $0$ is a $n$ by $n$ matrix of zeroes,
$I^R$ is the 
\nina{reflection} of the identity matrix,
and $P$ is any permutation matrix (a matrix with 
one non-zero entry on each row and column). 
For example, as there are exactly two possible
permutation matrices of order 2, there
are two symmetric 4 by 4 solutions with
lexicographically
ordered rows and columns:
\begin{eqnarray*}
\begin{array}{cccc}
0 & 0 & 0 & 1 \\
0 & 0 & 1 & 0 \\
0 & 1 & 1 & 0 \\
1 & 0 & 0 & 1 
\end{array}
& \ \ \ {\rm and} \ \ \ &
\begin{array}{cccc}
0 & 0 & 0 & 1 \\
0 & 0 & 1 & 0 \\
0 & 1 & 0 & 1 \\
1 & 0 & 1 & 0 
\end{array}
\end{eqnarray*}
In general, there are $n!$ row and column symmetries
of $P$. Hence, \DLex leaves $n!$ symmetric
solutions. 
\myqed

Having decided to break
row and column symmetry with
\DLex , how do we propagate it?
One option is to decompose it into two
$\LEXCHAIN$ constraints, one on the rows
and the other on the columns. A $\LEXCHAIN$ constraint
ensures that a sequence of vectors are lexicographically ordered. 
Enforcing domain consistency 
on each $\LEXCHAIN$ constraint takes polynomial time
\cite{lexchain}. However, 
this decomposition hinders propagation. 
For example, in the matrix of decision variables 
with domains:
$$
\begin{array}{ccc}
0/1 & 0/1 & 1 \\
0/1 & 0 & 1   \\
1 & 1 & 1
\end{array}
$$
\LEXCHAIN\ constraints on the
rows and columns ensure the second row is
lexicographically larger than
the first row and lexicographically smaller
than the third, and the second
column is lexicographically larger than
the first column and lexicographically
smaller than the third. 
Both such \LEXCHAIN\ constraints
are DC. However, 
the corresponding $\DLex$
constraint is not since there is no
solution in which the top left variable
is set to 1. 
We might 
therefore consider
a specialized propagator
for the \DLex constraint. 
Unfortunately, whilst checking a \DLex constraint takes
polynomial time, 
enforcing DC on this constraint is 
NP-hard. Thus, even when posting just \DLex to
break row and column symmetry, there are
computational limits on our ability to prune
symmetric branches from the search tree.

\begin{mytheorem}
\label{t:lex_double}
Enforcing DC on the  $\DLex$ constraint is NP-hard.
\end{mytheorem}
\myproof (Outline)
We reduce an instance of 1-in-3SAT on positive clauses
to a partially instantiated instance of the \DLex constraint
with 0/1 variables.
The constructed 
$\DLex$ constraint has a solution iff the 1-in-3 SAT formula is 
satisfiable. 
Hence, it is NP-hard to enforce DC on
the $\DLex$ constraint~\cite{bhhwaaai2004},
even with a bounded number of values. 
The full proof appears in~\cite{dlexproof}.
\myqed

\myOmit{
This answers a challenge proposed by
Frisch \textit{and al.} 
eight years ago:
\begin{quote}
{\em ``Global constraints for 
lexicographic orderings simultaneously
along both rows and columns of a matrix would also present a
significant challenge.''} (page 108 of \cite{fhkmwcp2002})
\end{quote}
We have shown that enforcing domain 
consistency on such a constraint is NP-hard. 
Our proof uses 0/1 variables so even 
enforcing bound consistency is NP-hard. }

\section{Special cases}

We consider two
special cases where we can check 
a constraint that breaks all row
and column symmetry in polynomial time. 
In both cases, we show that we 
can do even better than check
the constraint in polynomial time.
We prove that in these cases we can enforce DC
on a constraint that breaks all row
and column symmetry in polynomial time. 
This provides a counterpoint to our
result that enforcing DC
on \DLex is NP-hard in general. 
\vspace{-9pt}
\subsection{All-different matrices}

An all-different matrix is a matrix model
in which every value is different.
It was shown in~\cite{ffhkmpwcp2002}
that when an all-different matrix has row and column symmetry,
then $\RowColSym$ is equivalent to ensuring
that the top left entry is the smallest value,
and the first row and column are ordered.
Let
\OrderRowCol\ be such a symmetry breaking
constraint.

\begin{mytheorem}
DC can be enforced on
$\OrderRowCol$ in polynomial time. 
\end{mytheorem}
\myproof
Consider the $n$ by $m$ matrix model $X_{i,j}$. 
We post $O(nm)$ constraints:
$X_{1,1} < \ldots < X_{n,1}$,
$X_{1,1} < \ldots < X_{1,m}$,
$X_{1,1} < X_{1+i,1+j}$ for $1 \leq i < n$ and $1 \leq j < m$. 
The constraint graph of this decomposition
is acyclic. Therefore enforcing DC on the
decomposition achieves DC on $\OrderRowCol$. 
Each constraint in the decomposition
can be made DC in constant time (assuming
we can change bounds in constant time).
Hence, DC  can be enforced on $\OrderRowCol$
in $O(nm)$ time. 
\myqed

Note that, 
when applied to an all-different
matrix with row and column symmetry,
the general method for breaking 
symmetry in all-different
problems proposed in \cite{pijcai2005}
will post binary inequalities logically
equivalent to $\OrderRowCol$.

\subsection{Matrix models of functions}

A matrix model of a function is one in which
all entries are 0/1 and each row sum is 1. 
If a matrix model of a function has row and column symmetry
then $\RowColSym$ ensures 
the rows and columns are lexicographically
ordered, the row sums are 1, 
and the sums of the columns are in decreasing
order, as was shown in \cite{ilya01,lex2001,ffhkmpwcp2002}.
We denote this symmetry breaking constraint as $\DLexColSum$.
Enforcing DC on
$\DLexColSum$ takes polynomial time, in contrast
to partial row and column interchangeability
in matrix models of functions, which
is NP-hard~\cite{wcp07}. 

\begin{mytheorem}
DC can be enforced on $\DLexColSum$ in
polynomial time. 
\end{mytheorem}
\myproof
\nina{We will show that $\DLexColSum$ can be encoded
with a set of $\Regular$ constraints.
Consider the $n$ by $m$ matrix model $X_{i,j}$. 
For each row $i$ we introduce an extra variable $Y_i$
and a $\Regular$ constraint on $[X_{i,1}, \ldots,X_{i,m}, \#, Y_i]$
where $\#$ is a delimiter between
$X_{i,m}$ and $Y_i$. Each $\Regular$ 
constraint ensures that exactly one position in the $i$th
row is set to $1$ and the variable $Y_i$ stores this position.
The automaton's states are represented by the 3-tuple
$\left\langle s, d, p \right\rangle$
where $s$ is the row sum,
$d$ is the current position
and $p$ records the position of
the 1 on this row. 
This automaton has $4m$ states and a 
constant number of transitions from each state, so
the total number of transitions is $O(m)$. 
The complexity of propagating this constraint is
$O(m^2)$.
}
\nina{ We also post a $\Regular$ 
constraint over $Y_1,\ldots,Y_n$ to
ensure that they form a decreasing sequence of numbers
and the number of occurrences of each value is 
decreasing. 
The first condition ensures that
rows and columns are lexicographically
ordered and the second condition ensures that 
the sums of the columns are decreasing.
The states of this automaton are 3-tuples
$\left\langle v, s, r\right\rangle$ where $v$
is the last value, $s$ is the number of occurrences
of this value, and $r$ is the number of occurrences
of the previous value. 
%
%
This automaton has $O(n^2m)$ states,
while the number of transition from each state is bounded.
Therefore propagating this constraint requires time
$O(n^3m)$.
}
This decomposition is logically equivalent to
the $\DLexColSum$ constraint, therefore it is sound.
Completeness follows from the fact that
the decomposition has 
a Berge acyclic constraint graph. Therefore, 
enforcing DC on each $\Regular$
constraint enforces DC on $\DLexColSum$ in
$O(m^2n + n^3m)$ time.
\myOmit{We will encode $\DLexColSum$ 
with a $\Regular$ constraint. $\DLexColSum$ ensures
we have a line of 1s, starting in the top right
corner of the matrix and moving downwards and leftwards. 
Each vertical line of 1s is longer than the last. 
The $\Regular$ constraint is posted
on $X_{1,m}$ to $X_{1,1}$, then
$X_{2,m}$ to $X_{2,1}$, and so
on up to $X_{n,1}$.
} 
\myOmit{
The automaton defining the regular
language has states which
are 5-tuples of the form $(sum,column,last,oldsum,newsum)$
where $sum$ is the sum of the current row,
$column$ is the current 
column position, 
$last$ is the column of the last 1 to appear, 
$oldsum$ is the number of 1s in column $last$,
and $newsum$ is the number of 1s so far in column $last$. 
The automaton starts in the state $(0,m,0,0,0)$.
From this state we only accept the value 1 (for $X_{1,m}$)
and move to the state $(1,m-1,m,0,1)$. 
If we are in the state $(1,i,last,oldsum,newsum)$ 
then we accept only the value 0 and move to the state
$(1,i-1,last,oldsum,newsum)$ for $i>1$ 
and the state
$(0,m,last,oldsum,newsum)$ for $i=1$.
If we are in the state $(0,i,last,oldsum,newsum)$ and $i>last$
then we only accept the value
0 and move to the state $(0,i-1,last,oldsum,newsum)$.
If we are in the state $(0,last,last,oldsum,newsum)$ 
and $newsum < oldsum$ then we only accept the value
1 and move to the state $(1,last-1,last,oldsum,newsum+1)$ 
if $last>1$ and to the state $(0,m,1,oldsum,newsum+1)$ 
if $last=1$. 
If we are in the state $(0,last,last,oldsum,newsum)$ 
then we accept the value
1 and move to the state $(1,last-1,last,oldsum,newsum+1)$ 
if $last>1$ and to the state $(0,m,1,oldsum,newsum+1)$ 
if $last=1$.
If we are in the state $(0,last,last,oldsum,newsum)$, $last>1$
and $newsum \geq oldsum$ then we accept the value
0 and move to the state $(0,last-1,last,oldsum,newsum)$.
Finally, if we are in the state $(0,last-1,last,oldsum,newsum)$
and $last>1$ then we accept the value 1 and move to the
state 
$(1,last-2,last-1,newsum,1)$ for $last>2$,
and to the state 
$(0,m,1,newsum,1)$ for $last=2$.
The only accepting state is of the form $(0,m,last,oldsum,newsum)$
where $newsum \geq oldsum$. }
\myOmit{
The automaton defining the regular
language has states which
are of the form $(s,c,l,o,n)$
where $s$ is the sum of the current row,
$c$ is the current 
column position, 
$l$ is the column of the last 1 to appear, 
$o$ is the number of 1s in column $l$,
and $n$ is the number of 1s so far in column $l$. 
The automaton starts in the state $(0,m,0,0,0)$.
The only accepting state is of the form $(0,m,l,o,n)$
where $n \geq o$.
Transitions only permit a 1 to
appear in the current column till
the column sum equals the previous
one, and then only permit
1 to occur in the current column or
the next earlier. 
Since column sums are bounded by $n$,
there are $O(n^2m^2)$ states. Hence, we can
enforce DC on the $O(nm)$ variables in
$O(n^3m^3)$ time. 
}
\myqed

\section{Value symmetry}

Problems with row and column symmetry also often
contain value symmetries. For example, 
the EFPA problem has row, column and
value symmetry.
We therefore turn to the problem of breaking
row, column and value symmetry. 

\begin{myexample}
Consider again the solution 
\mbox{\rm (\ref{epfa})}.
If we interchange the values
1 and 2, we get
a symmetric solution:
\renewcommand{\theequation}{\alph{equation}}
\begin{equation} \label{epfa4}
\begin{array}{cccccc}
0 & 2 & 1 & 2 & 0 & 1 \\
0 & 2 & 2 & 1 & 1 & 0 \\
0 & 1 & 0 & 2 & 1 & 2 \\
0 & 0 & 1 & 1 & 2 & 2  
\end{array}
\
\begin{array}{c}
\ \ \ \ \ \ \Rightarrow \ \ \ \ \ \ \\
\mbox{\rm (1 2)}
\end{array}
\ \
\begin{array}{cccccc}
0 & 1 & 2 & 1 & 0 & 2 \\
0 & 1 & 1 & 2 & 2 & 0 \\
0 & 2 & 0 & 1 & 2 & 1 \\
0 & 0 & 2 & 2 & 1 & 1
\end{array}
\end{equation}
In fact, all values in
this CSP are interchangeable. 
\end{myexample}

How do we break value symmetry
in addition to breaking row
and column symmetry? For example,
Huczynska {\it et al.}
write about their first model of the EFPA problem:
\begin{quote}
{\em ``To break {some} of the
symmetry, we apply lexicographic ordering (lex-ordering) constraints to the rows and
columns \ldots
These two constraint sets do not explicitly order the symbols. It would be possible
to order the symbols by using value symmetry breaking constraints. 
However we leave this for {future} work.''} (page 53 of \cite{hmmncp09})
\end{quote}
We turn to this future work of breaking row, column
and value symmetry. 

\subsection{Double Lex}

We first note that 
the interaction of the problem and \DLex 
constraints can in some circumstances
break all value symmetry. 
For instance, in our (and 
Huczynska {\it et al.}'s) model of the EFPA problem,
\emph{all} value symmetry is already eliminated. This appears
to have been missed by
\cite{hmmncp09}. 

\begin{myexample}
Consider any solution 
of the EFPA problem
which satisfies \DLex
\renewcommand{\theequation}{\alph{equation}}
(e.g. \mbox{\rm (\ref{epfa2})}
or \mbox{\rm (\ref{epfa3})}).
By ordering columns
lexicographically, \DLex ensures
that the first row is ordered. 
In addition, the problem constraints
ensure  $\lambda$ copies of the symbols
1 to $q$ to appear in the first row. Hence, 
the first row is forced to be:
$$
\overbrace{1\ldots 1}^{\lambda}
\overbrace{2\ldots 2}^{\lambda}
\ldots
\overbrace{q\ldots q}^{\lambda}
$$
All value symmetry is
broken as we cannot permute the occurrences
of any of the values. 
\end{myexample}

\subsection{Puget's method}

In general, value
symmetries may remain
after we have broken row and column symmetry. 
How can we eliminate these value
symmetries?
Puget has given a
general method for breaking any number of 
value symmetries in polynomial time \cite{pcp05}. 
Given a surjection problem in which all values
occur at least once,\footnote{Any problem can
be turned into a surjection problem by the
addition of suitable new variables.} he introduces
variables $Z_j$ to represent the index
of the first occurrence of each value:
\begin{eqnarray*}
X_i = j  & \Rightarrow & Z_j \leq i \\
Z_j = i & \Rightarrow & X_i=j
\end{eqnarray*}
Value symmetry on the $X_i$ is transformed
into variable symmetry on the $Z_j$. 
This variable symmetry is especially 
easy to break as the $Z_j$ take all different
values. We simply need to post appropriate
ordering constraints on the $Z_j$. 
Consider, for example, the inversion
symmetry which maps $1$ onto $m$,
$2$ onto $m-1$, etc.
Puget's method breaks this symmetry
with the single ordering constraint:
$Z_1  <  Z_m$.
Unfortunately Puget's method for breaking
value symmetry is not
compatible in general with breaking
row and column symmetry using \RowColSym .
This corrects Theorem 6 and Corollary 7 in \cite{pcp05}
which claim that, provided
we use the same ordering of variables in each method,
it is compatible to post lex-leader constraints to break
variable symmetry and Puget's constraints
to break value symmetry. There is no ordering
of variables in Puget's method which is 
compatible with breaking row and column symmetry
using the lex-leader method (or any method
like \DLex based on it). 

\begin{mytheorem}
There exist problems on
which posting \RowColSym\ and applying Puget's method for
breaking value symmetry remove all solutions
in a symmetry class irrespective of the ordering
on variables used by Puget's method.
\end{mytheorem}
\myproof
Consider a 3 by 3 matrix model with 
constraints that all values between
0 and 8 occur, and 
that the average of the non-zero values
along every row and column
are all different from
each other. This problem has row and column
symmetry since we can permute any pair
of rows or columns without changing the
average of the non-zero values.
In addition, it has a value
symmetry that maps $i$ onto $9-i$ for $i>0$.
This maps an average of $a$ onto $9-a$.
If the averages were all-different before
they remain so after. 
Consider the following two solutions:
\begin{eqnarray*}
\begin{array}{ccc}
0 & 2 & 3 \\
4 & 8 & 5 \\
7 & 6 & 1 
\end{array} 
&
\ \ \ {\rm and} \ \ \ 
&
\begin{array}{ccc}
0 & 2 & 3 \\
4 & 1 & 5 \\
7 & 6 & 8 
\end{array} 
\end{eqnarray*}
Both matrices satisfy \RowColSym\ as the
smallest entry occurs in the top left
corner and both the first row and column
are ordered. They are therefore both
the lex leader members of their
symmetry class. 

Puget's method 
for breaking value symmetry will simply
ensure that the first occurrence of 1 
in some ordering of the matrix is before that of 8
in the same ordering. However, comparing
the two solutions, it cannot
be the case that the middle square is
both before {\em and} after the bottom right
square in the given ordering used by Puget's method. 
Hence, whichever ordering
of variables is used by Puget's method,
one of these solutions will be eliminated.
All solutions in this symmetry class are
thus eliminated. 
\myqed


We can pinpoint the 
mistake in Puget's proof which
allows him to conclude incorrectly that
his method for value symmetry can be safely combined with
variable symmetry breaking methods like \DLex . 
Puget introduces a matrix of 0/1 variables $Y_{ij} \iff X_i = j$
and observes that variable symmetries $\sigma$ on variables $X_i$
correspond to row
symmetries on the matrix $Y_{ij}$, while value symmetries $\theta$ of
the variables $X_i$ correspond to column symmetries of the matrix.
Using the lex-leader method on
a column-wise linearisation of the matrix, 
he derives the value symmetry breaking constraints on the $Z$ variables. 
Finally, he claims
that we can derive the variable symmetry
breaking constraints on the $X$ variables
with the same method
(equation (13) of \cite{pcp05}).
However, this requires a row-wise
linearisation of the matrix. 
Unfortunately, combining symmetry breaking
constraints based on row and column-wise
linearisations can, as in our example,
eliminate all solutions in a symmetry class. 


%
%
In fact, we can give an even
stronger counter-example to Theorem 6 in \cite{pcp05}
which shows that
it is incompatible to post together
variable and value symmetry breaking constraints
{\em irrespective} of the orderings of variables used
by {\em both} the variable and the value
symmetry breaking method. 

\begin{mytheorem}
There exist problems on
which posting lex-leader constraints
to break variable symmetries and applying Puget's method 
to break value symmetries remove all solutions
in a symmetry class irrespective of the orderings
on variables used by both methods.
\end{mytheorem}
\myproof
Consider variables $X_1$ to $X_4$ taking values 1 to 4,
an all-different constraint over $X_1$ to $X_4$ and
a constraint that the neighbouring differences
are either all equal or 
are not an arithmetic sequence. 
These constraints permit 
solutions like $X_1,\ldots,X_4=1,2,3,4$ 
(neighbouring differences are all equal)
and $X_1,\ldots,X_4=2,1,4,3$ 
(neighbouring differences are not
an arithmetic sequence). They
rule out assignments like
$X_1,\ldots,X_4=3,2,4,1$ 
(neighbouring differences form
the arithmetic sequence $1, 2, 3$).
This problem has a variable symmetry $\sigma$ which
reflects a solution, swapping $X_1$ with $X_4$,
and $X_2$ with $X_3$, and a value
symmetry $\theta$ that inverts
a solution, swapping $1$ with $4$, and
$2$ with $3$. 
Consider 
$X_1,\ldots,X_4=2,4,1,3$
and 
$X_1,\ldots,X_4=3,1,4,2$.
These two assignments 
form a symmetry class 
of solutions. 

Suppose we break variable symmetry with
a lex-leader constraint on $X_1$ to $X_4$.
This will permit the solution 
$X_1,\ldots,X_4=2,4,1,3$
and eliminate the solution
$X_1,\ldots,X_4=3,1,4,2$.
Suppose we break the value 
symmetry using Puget's method on
the same ordering of variables. This
will ensure that $1$ first occurs before
$4$. But this will eliminate the
solution 
$X_1,\ldots,X_4=2,4,1,3$. 
Hence, all solutions in this 
symmetry class are eliminated.
In this case, both variable and value
symmetry breaking use the same
order on variables. However, we
can show that all solutions in at least
one symmetry class are eliminated
whatever the orders used by both the variable
and value symmetry breaking.


The proof is by case analysis. In each case,
we consider a set of symmetry classes of solutions,
and show that the combination
of the lex-leader constraints to break
variable symmetries and Puget's method
to break value symmetries 
eliminates all solutions
from one symmetry class.
In the first case, suppose the variable
and value symmetry breaking constraints
eliminate $X_1,\ldots,X_4=3,1,4,2$
and permit 
$X_1,\ldots,X_4=2,4,1,3$. 
In the second case, suppose they
eliminate $X_1,\ldots,X_4=2,4,1,3$
and permit 
$X_1,\ldots,X_4=3,1,4,2$. 
This case is symmetric to 
the first except we need to
reverse the names of the variables 
throughout the proof. We therefore
consider just the first case. In this
case, the lex-leader constraint breaks the 
variable symmetry by putting either $X_1$ first 
in its ordering variables or $X_3$ first. 

Suppose $X_1$ goes first in the ordering 
used by the lex-leader constraint.
Puget's method 
ensures that the first occurrence of 1 is before that of
4. Puget's method therefore uses
an ordering on variables which puts
$X_3$ before $X_2$. 
Consider now the symmetry class of 
solutions: $X_1,\ldots,X_4=2,1,4,3$
and $X_1,\ldots,X_4=3,4,1,2$. 
Puget's
method eliminates the first
solution as 4 occurs before 1 in any
ordering that put $X_3$ before $X_2$. 
And the lex-leader constraint eliminates
the second solution as $X_1$ is larger
than its symmetry $X_4$.  Therefore
all solutions in this symmetry class are eliminated.

Suppose, on the other hand, $X_3$ goes
first in the lex-leader constraint. 
Consider now the symmetry class of 
solutions: $X_1,\ldots,X_4=1,2,3,4$
and $X_1,\ldots,X_4=4,3,2,1$. 
The lex-leader constraint eliminates
the first solution as $X_3$ is greater
than its symmetry $X_2$. Suppose now that the
second solution is not eliminated. 
Puget's method ensures the first occurrence of 1 is before that of
4. Puget's method therefore uses
an ordering on variables which puts
$X_4$ before $X_1$. 
Consider now the symmetry class of 
solutions: $X_1,\ldots,X_4=1,3,2,4$
and $X_1,\ldots,X_4=4,2,3,1$. 
Puget's
method eliminates the first
solution as 4 occurs before 1 in any
ordering that put $X_4$ before $X_1$. 
And the lex-leader constraint eliminates
the second solution as $X_3$ is larger
than its symmetry $X_2$.  Therefore
all solutions in this symmetry class are eliminated.
\myqed

\subsection{Value precedence}

We end with a special but common case where variable and 
value symmetry breaking do not conflict.
When values partition into interchangeable sets, 
Puget's method is equivalent to breaking
symmetry by enforcing value precedence \cite{llcp2004,wecai2006}.
Given any two interchangeable
values $i$ and $j$ with $i<j$, a value
\precedence\ constraint ensures that if $i$ occurs
then the first occurrence of $i$ is before
that of $j$. It is safe to break row and column symmetry
with \RowColSym\ and value symmetry
with \precedence\ when value precedence considers
variables either in a row-wise or in a column-wise order. 
This is a simple consequence of Theorem 1 in \cite{llcp2004}.
It follows that it is
also safe to use \precedence\ to break
value symmetry when 
using constraints like \DLex derivable
from the lex-leader method. 

\section{Snake Lex}

A promising alternative to \DLex for 
breaking row and column symmetries 
is \snakelex \cite{snakelex}. This is 
also derived from the lex leader method, but
now applied to a snake-wise unfolding of the matrix.
To break column symmetry, \snakelex
ensures that the first column is 
lexicographically smaller than or equal to
both the second and third columns,
the reverse of the second column
is lexicographically smaller than or equal to
the reverse of both the third and fourth columns,
and so on up till the penultimate column
is compared to the final column.
To break row symmetry, \snakelex
ensures that each neighbouring pair
of rows, $X_{1,i},\ldots,X_{n,i}$
and $X_{1,i+1},\ldots,X_{n,i+1}$ satisfy the entwined
lexicographical ordering:
\begin{eqnarray*}
\langle X_{1,i}, X_{2,i+1}, X_{3,i}, X_{4,i+1}, \ldots \rangle & \leq_{\rm lex} 
\langle X_{1,i+1}, X_{2,i}, X_{3,i+1}, X_{4,i}, \ldots \rangle
\end{eqnarray*}

\myOmit{
\DLex breaks the subset of the row and column
symmetries that swap neighbouring rows and columns.
\snakelex , by comparison, breaks a strict 
superset of these symmetries.
%
To see this, 
consider the subset of \snakelex
constraints that ensure the first column is 
lexicographically smaller than or equal to
the second column,
the reverse of the second column is lexicographically
smaller than or equal to
the reverse of the third, and so on
up to the ordering of the final
two columns, and the row constraints
that ensure neighbouring rows satisfy the entwined
lexicographical ordering. The column
constraints can be derived from the lex leader
constraints for the symmetries
which swap neighbouring columns,
taking the matrix in snake order
and deleting those parts of the lexicographical
ordering constraint where the vectors are 
equal. 
Similarly, the row constraints
constraints can be derived from the lex leader
constraints for the symmetries
which swap neighbouring rows, again
taking the matrix in snake order
and deleting those parts of the lexicographical
ordering constraint where the vectors are 
equal. 

To see that \snakelex breaks strictly more symmetry, consider 
the following two row and column symmetric matrices:
\begin{eqnarray*}
\begin{array}{ccc}
0&1&0 \\
0&1&1 \\
1&0&1
\end{array}
& \ \  {\rm and} \ \  & 
\begin{array}{ccc}
0&1&0 \\
1&1&0 \\
1&0&1
\end{array}
\end{eqnarray*}
The second matrix can be obtained from
the first by interchanging the first and third
columns. 
However, only the first satisfies the
\snakelex constraints. The second
matrix satisfies the subset of the 
\snakelex constraints that 
prevent neighbouring columns
being interchanged. However,
the second matrix violates the 
\snakelex constraint that the first
column is lexicographically smaller
than or equal to the third. 
%
}
Like \DLex, \snakelex is an incomplete
symmetry breaking method. 
In fact, like \DLex , it may leave
a large number of symmetric solutions. 

\begin{mytheorem}
\label{tm:snake-expsol}
There exists a class of $2n$ by $2n+1$
0/1 matrix models on which
\snakelex leaves $O(4^n/\sqrt{n})$ symmetric solutions,
for all $n \geq 2$.
\end{mytheorem}
\myproof
Consider the following 4 by 4 matrix:
\begin{eqnarray*}
&
\begin{array}{cccc} 
0 & 1 & 0 & 0 \\
0 & 0 & 0 & 1 \\
0 & 0 & 1 & 0 \\
1 & 0 & 0 & 0
\end{array}
& 
\end{eqnarray*}
This is a permutation matrix  as there is a single 
1 on each row and column. 
It satisfies the \snakelex constraints.
In fact, we can add any 5th column
which reading top to bottom
is lexicographically larger than or
equal to $0010$ and reading bottom
to top is lexicographically larger than 
or equal to $0010$. We shall add 
a 4 bit column with 2 bits set.
That is, reading top to bottom:
$1100$,  $1010$, $0110$ or $0011$.
Note that  all 4 of these 4 by 5 matrices
are row and column symmetries of each
other. For instance, consider the 
row and column symmetry
$\sigma$ that reflects the matrix in
the horizontal axis, and swaps
the 1st column with the 2nd,
and the 3rd with the 4th:
\begin{eqnarray*}
\begin{array}{ccccc} 
0 & 1 & 0 & 0 & 1\\
0 & 0 & 0 & 1 & 1\\
0 & 0 & 1 & 0 & 0\\
1 & 0 & 0 & 0 & 0
\end{array}
&
\begin{array}{c} \\
\ \ \ \ \Leftrightarrow \ \ \ \ \\
\sigma
\end{array}
& 
\begin{array}{ccccc} 
0 & 1 & 0 & 0 & 0\\
0 & 0 & 0 & 1 & 0\\
0 & 0 & 1 & 0 & 1\\
1 & 0 & 0 & 0 & 1
\end{array}
\end{eqnarray*}
In general, we consider the $2n$ by $2n$ permutation matrix:
\begin{eqnarray*}
&
{\small
\begin{array}{cccccccccc} 
0 & 1 & 0 & 0 & 0 & 0 & \ldots & 0 & 0 & 0\\
0 & 0 & 0 & 1 & 0 & 0 & \ldots & 0 & 0 & 0\\
0 & 0 & 0 & 0 & 0 & 0 & \ldots & 0 & 0 & 0\\
0 & 0 & 0 & 0 & 0 & 1 & \ldots & 0 & 0 & 0\\
\vdots & \vdots & \vdots & \vdots & \vdots & \vdots &  & \vdots & \vdots & \vdots \\
0 & 0 & 0 & 0 & 0 & 0 & \ldots & 1 & 0 & 0 \\
0 & 0 & 0 & 0 & 0 & 0 & \ldots & 0 & 0 & 1 \\
0 & 0 & 0 & 0 & 0 & 0 & \ldots & 0 & 1 & 0 \\
\vdots & \vdots & \vdots & \vdots & \vdots & \vdots &  & \vdots & \vdots & \vdots \\
0 & 0 & 0 & 0 & 1 & 0 & \ldots & 0 & 0 & 0 \\
0 & 0 & 1 & 0 & 0 & 0 & \ldots & 0 & 0 & 0 \\
1 & 0 & 0 & 0 & 0 & 0 & \ldots & 0 & 0 & 0 
\end{array}
}
& 
\end{eqnarray*}
This 
satisfies the \snakelex constraints.
We can add any $2n+1$th column
which reading top to bottom is
lexicographically larger than or equal to the $2n-1$th 
column and reading bottom to
top is lexicographically larger than 
or equal to the $2n$th column. In fact,
we can add any column with eactly $n$ of the 
$2n$ bits set. This gives us a set of
$2n$ by $2n+1$ matrices
that are row and column symmetries of
each other. 
There are $(2n)!/(n!)^2$ bit vectors with
exactly $n$ of $2n$ bits set. 
Hence, we have $(2n)!/(n!)^2$ matrices
which satisfy \snakelex that are in the
same row and column symmetry class. 
Using Stirling's formula, this grows as 
$O(4^n/\sqrt{n})$. 
\myqed

\section{Experimental results}

%

The proof of Theorem~\ref{tm:fpt} gives
a polynomial method to break all row and
column symmetry. 
This allows us to compare symmetry breaking methods
for matrix models like \DLex and \snakelex, not only
with respect to each other but for the first time in absolute terms.
Our aim is to evaluate: first,
whether the worst-case scenarios identified in 
theorems~\ref{tm:expsol} and~\ref{tm:snake-expsol}
are indicative of what can be expected in practice; 
second, how effective these methods are with respect 
to each other; third, in cases where they differ
significantly,
how much closer the best of them is to the optimal.


To answer these questions,
we experimented with different symmetry breaking constraints:
$\DLex$, the column-wise \snakelex ($\SnakeLex_C$)  or the
row-wise \snakelex ($\SnakeLex_R$) \cite{snakelex}.  
We use $\NoSymShort$ to denote no symmetry breaking constraints. 
For each problem instance we found the total number of solutions left by
symmetry breaking constraints ($ \#s$)  and  computed how many of them were symmetric based 
on the method outlined in the proof of Theorem~\ref{tm:fpt}. 
The number of \emph{non symmetric} solutions
is equal to the number of symmetry classes ($\#ns$) if 
the search space is exhausted. In all instances
at least one model exhausted the search space
to compute the of symmetry classes,
shown in the column $\RowColSymShort$.
We use `$-$' to  indicate that the search is not 
completed within the time limit.
As the \NoSymShort\ model typically could not exhaust
the search space within the time limit, we use `$>$' to indicate
a lower bound on the number of solutions.
Finally, we used a variable ordering heuristic that follows the 
corresponding lex-leader  variable ordering 
in each set of symmetry breaking constraints (i.e.
row-wise snake ordering with 
$\SnakeLex_R$).
We ran experiments in Gecode 3.3.0 
on an Intel XEON X5550, 2.66 GHz, 
32 GB RAM with $18000$ sec timeout. 


\emph{Unconstrained problems.}
We first evaluated the effectiveness of 
symmetry breaking constraints 
in the absence of problem constraints. This
gives the ``pure'' effect of these constraints
at eliminating row and column symmetry. 
We considered a problem with a matrix  $m_{r \times c}$, $r \leq c = [2,6]$,
$D(m_{r,c}) = [0,d-1]$, $d=[2,\ldots,5]$ whose rows and columns are interchangeable.
Table~\ref{t:t2} summarizes the results. The first part
presents typical results for 0/1 matrices whilst
the second part presents results for larger domains. 
The results support 
the exponential worst case in
Theorems~\ref{tm:expsol} and~\ref{tm:snake-expsol},
as the ratio of solutions found to symmetry classes
increases from 1.25 (3,3,2) to over 6 (6,6,2), 
approximately doubling with each increase of 
the matrix size.
As we increase the problem size, 
the number of symmetric solutions left 
by \DLex and \SnakeLex\ grows rapidly.
Interestingly, $\SnakeLex_C$ achieves better pruning on 
0/1 matrices,  while $\DLex$ performs better with
larger domains.

\begin{table}
\begin{center}
{\scriptsize
\caption{\label{t:t2} Unconstrained problems. 
Number of solutions found by posting different sets of symmetry breaking constraints. 
$r$ is the number of rows, $c$ is the number of columns, 
$d$ is the size of the domains. 
}
\begin{tabular}{|r||c|c|c|c|c|c|}
\hline
$(r, c, d )$& {$\RowColSymShort$}
&{$\NoSymShort$}
&{$\DLex$}
&{$\SnakeLex_R$}
&{$\SnakeLex_C$}\\
\hline 
\hline
 & 
 \#ns&
 \#s &
 \#s / time& 
 \#s / time& 
 \#s / time\\ 
\hline 
$ (3,3,2) $  &  $36$ &  $ 512 $    &     $ 45 $  / $ \mathbf{0.00} $  &     $ \mathbf{44} $  / $ \mathbf{0.00} $  &     $ \mathbf{44} $  / $ \mathbf{0.00} $ \\
$ (4,4,2) $  &  $317$ &  $ 65536 $    &     $ 650 $  / $ \mathbf{0.00} $  &     $ \mathbf{577} $  / $ \mathbf{0.00} $  &     $ \mathbf{577} $  / $ \mathbf{0.00} $ \\
$ (5,5,2) $  &  $5624$ &  $ 3.36{\cdot} 10^7 $    &     $ 24520 $  / $ \mathbf{0.05} $  &     $ \mathbf{18783} $  / $ 0.06 $  &     $ \mathbf{18783} $  / $ 0.06 $ \\
$ (6,6,2) $ & $251610$ & $ > 9.4{\cdot} 10^9$ & $2.62 \cdot 10^6$ / $ 22.2 $ & $ \mathbf{1.71 \cdot 10^6}$  / $ 22.2 $ & $ \mathbf{1.71 \cdot 10^6}$ / $\mathbf{18.1}$\\
\hline 
\hline 
$ (3,3,3) $  &  $738$ &  $ 19683 $    &     $ \mathbf{1169} $  / $ \mathbf{0.00} $  &     $ 1232 $  / $ \mathbf{0.00} $  &     $ 1232 $  / $ \mathbf{0.00} $ \\
$ (3,3,4) $  &  $8240$ &  $ 2.62{\cdot} 10^5 $    &     $ \mathbf{14178} $  / $ 0.03 $  &     $ 15172 $  / $ \mathbf{0.02} $  &     $ 15172 $  / $ 0.05 $ \\
$ (3,3,5) $  &  $57675$ &  $ 1.95{\cdot} 10^6 $    &     $ \mathbf{1.02{\cdot} 10^5} $  / $ 0.19 $  &     $ 1.09{\cdot} 10^5 $  / $ \mathbf{0.15} $  &     $ 1.09{\cdot} 10^5 $  / $ 0.21 $ \\
$ (3,3,6) $  &  $289716$ &  $ 1.01{\cdot} 10^7 $    &     $ \mathbf{5.20{\cdot} 10^5} $  / $ \mathbf{2.32} $  &     $ 5.54{\cdot} 10^5 $  / $ 3.29 $  &     $ 5.54{\cdot} 10^5 $  / $ 2.83 $ \\
\hline 
\end{tabular}}
\end{center}
%
\begin{center}
{\scriptsize
\caption{\label{t:t1} Equidistant Frequency Permutation Array problems. 
Number of solutions found by posting different sets of symmetry breaking constraints. 
$v$ is the number code words, $q$ is the number of different symbols, 
$\lambda$ is the size of the domains. }
\begin{tabular}{|r||c|c|c|c|c|c|}
\hline
$(q, \lambda, d, v)$& {$\RowColSymShort$}
&{$\NoSymShort$}
&{$\DLex$}
&{$\SnakeLex_R$}
&{$\SnakeLex_C$}\\
\hline 
\hline
 & 
 \#ns&
 \#s &
 \#s / time& 
 \#s / time& 
 \#s / time\\ 
\hline 
$ (3, 3,2,3) $  &  $6$ &  $ 1.81{\cdot} 10^5 $    &     $ \mathbf{6} $  / $ \mathbf{0.00} $  &     $ \mathbf{6} $  / $ \mathbf{0.00} $  &     $ \mathbf{6} $  / $ \mathbf{0.00} $ \\
$ (4, 3,3,3) $  &  $8$ &   $ > 3.88{\cdot} 10^7 $    &     $ \mathbf{16} $  / $ \mathbf{0.01} $  &     $ \mathbf{16} $  / $ 0.01 $  &     $ \mathbf{16} $  / $ 0.16 $ \\
$ (4, 4,2,3) $  &  $12$ &   $ > 5.87{\cdot} 10^7 $    &     $ \mathbf{12} $  / $ \mathbf{0.00} $  &     $ \mathbf{12} $  / $ \mathbf{0.00} $  &     $ \mathbf{12} $  / $ 0.04 $ \\
$ (3, 4,6,4) $  &  $1427$ &   $ > 5.57{\cdot} 10^7 $    &     $ 11215 $  / $ 5.88 $  &     $ 10760 $  / $ \mathbf{5.36} $  &     $ \mathbf{8997} $  / $ 493.87 $ \\
$ (4, 3,5,4) $  &  $8600$ &   $ > 2.03{\cdot} 10^7 $    &     $ 61258 $  / $ 69.90 $  &     $ 58575 $  / $ \mathbf{51.62} $  &     $ \mathbf{54920} $  / $ 3474.09 $ \\
$ (4, 4,5,4) $  &  $9696$ &   $ > 5.45{\cdot} 10^6 $    &     $ 72251 $  / $ 173.72 $  &     $ 66952 $  / $ \mathbf{132.46} $  &     $ \mathbf{66168} $  / $ 14374.82 $ \\
$ (5, 3,3,4) $  &  $5$ &   $ > 4.72{\cdot} 10^6 $    &     $ \mathbf{20} $  / $ 0.36 $  &     $ \mathbf{20} $  / $ \mathbf{0.25} $  &     $ \mathbf{20} $  / $ 31.61 $ \\
$ (3, 3,4,5) $  &  $18$ &   $ > 2.47{\cdot} 10^7 $    &     $ 71 $  / $ 0.17 $  &     $ 71 $  / $ \mathbf{0.13} $  &     $ \mathbf{63} $  / $ 30.08 $ \\
$ (3, 4,6,5) $  &  $4978$ &   $ > 2.08{\cdot} 10^7 $    &     $ 77535 $  / $ 167.50 $  &     $ \mathbf{71186} $  / $ \mathbf{137.88} $  & $-$  \\
$ (4, 3,4,5) $  &  $441$ &   $ > 6.55{\cdot} 10^6 $    &     $ 2694 $  / $ 19.37 $  &     $ 2688 $  / $ \mathbf{12.80} $  &     $ \mathbf{2302} $  / $ 5960.43 $ \\
$ (4, 4,2,5) $  &  $12$ &   $ > 6.94{\cdot} 10^6 $    &     $ \mathbf{12} $  / $ 0.02 $  &     $ \mathbf{12} $  / $ \mathbf{0.01} $  &     $ \mathbf{12} $  / $ 1.60 $ \\
$ (4, 4,4,5) $  &  $717$ &   $ > 6.27{\cdot} 10^6 $    &     $ 4604 $  / $ 38.15 $  &     $ \mathbf{4397} $  / $ \mathbf{24.58} $  & $-$  \\
$ (4, 6,4,5) $  &  $819$ &   $ > 4.08{\cdot} 10^6 $    &     $ 5048 $  / $ 69.83 $  &     $ \mathbf{4736} $  / $ \mathbf{44.83} $  & $-$  \\
$ (5, 3,4,5) $  &  $3067$ &   $ > 2.39{\cdot} 10^6 $    &     $ 20831 $  / $ 403.97 $  &     $ \mathbf{20322} $  / $ \mathbf{216.93} $  & $-$  \\
$ (6, 3,4,5) $  &  $15192$ &   $ > 2.16{\cdot} 10^6 $    &     $ 1.11{\cdot} 10^5 $  / $ 4924.41 $  &     $ \mathbf{1.06{\cdot} 10^5} $  / $ \mathbf{2006.19} $  & $-$  \\
\hline 
\end{tabular}}
\end{center}
%
%
%
\begin{center}
{\scriptsize
\caption{\label{t:t3} Balanced Incomplete Block Designs. 
Number of solutions found by posting different sets of symmetry breaking constraints. 
$v$ is the number of objects, $k$ is the objects in each block, 
every two distinct objects occur together in exactly $\lambda$ blocks. 
}
\begin{tabular}{|r||c|c|c|c|c|c|}
\hline
$(v, k, \lambda )$& {$\RowColSymShort$}
&{$\NoSymShort$}
&{$\DLex$}
&{$\SnakeLex_R$}
&{$\SnakeLex_C$}\\
\hline 
\hline
 & 
 \#ns&
 \#s &
 \#s / time& 
 \#s / time& 
 \#s / time\\ 
\hline 
$ (5,2,7) $  &  $1$ &  $ > 0 $    &     $ \mathbf{1} $  / $ \mathbf{0.01} $  &     $ \mathbf{1} $  / $ 0.02 $  &     $ \mathbf{1} $  / $ 73.26 $ \\
$ (5,3,6) $  &  $1$ &   $ > 1.51{\cdot} 10^9 $    &     $ \mathbf{1} $  / $ \mathbf{0.00} $  &     $ \mathbf{1} $  / $ \mathbf{0.00} $  &     $ \mathbf{1} $  / $ 0.82 $ \\
$ (6,3,4) $  &  $4$ &   $ > 1.29{\cdot} 10^9 $    &     $ \mathbf{21} $  / $ 0.01 $  &     $ 25 $  / $ \mathbf{0.00} $  &     $ \mathbf{21} $  / $ 12.62 $ \\
$ (6,3,6) $  &  $6$ &   $ > 1.21{\cdot} 10^9 $    &     $ \mathbf{134} $  / $ \mathbf{0.04} $  &     $ 146 $  / $ 0.07 $  &     $ \mathbf{134} $  / $ 1685.58 $ \\
$ (7,3,4) $  &  $35$ &   $ > 1.18{\cdot} 10^9 $    &     $ \mathbf{3209} $  / $ \mathbf{0.33} $  &     $ 9191 $  / $ 1.07 $  &     $ 5270 $  / $ 7241.92 $ \\
$ (7,3,5) $  &  $109$ &   $ > 1.09{\cdot} 10^9 $    &     $ 33304 $  / $ \mathbf{4.15} $  &     $ 85242 $  / $ 11.90 $  & $-$  \\
\hline 
\end{tabular}}
\end{center}
%
\begin{center}
{\scriptsize
\caption{\label{t:t4} Covering Arrays. 
Number of solutions found by posting different sets of symmetry breaking constraints. 
$b$ is the number of vectors, $k$ is the length of a vector, 
$g$ is the size of the domains, $t$ is the covering strength.}
\begin{tabular}{|r||c|c|c|c|c|c|}
\hline
$(t, k, g, b )$& {$\RowColSymShort$}
&{$\NoSymShort$}
&{$\DLex$}
&{$\SnakeLex_R$}
&{$\SnakeLex_C$}\\
\hline 
\hline
 & 
 \#ns&
 \#s &
 \#s / time& 
 \#s / time& 
 \#s / time\\ 
\hline 
$ (2, 3,2,4) $  &  $2$ &  $ 48 $    &     $ \mathbf{2} $  / $ \mathbf{0.00} $  &     $ \mathbf{2} $  / $ \mathbf{0.00} $  &     $ \mathbf{2} $  / $ \mathbf{0.00} $ \\
$ (2, 3,2,5) $  &  $8$ &  $ 1440 $    &     $ \mathbf{15} $  / $ \mathbf{0.00} $  &     $ \mathbf{15} $  / $ \mathbf{0.00} $  &     $ \mathbf{15} $  / $ \mathbf{0.00} $ \\
$ (2, 3,3,9) $  &  $6$ &  $ 4.35{\cdot} 10^6 $    &     $ \mathbf{12} $  / $ \mathbf{0.00} $  &     $ \mathbf{12} $  / $ \mathbf{0.00} $  &     $ \mathbf{12} $  / $ 1.95 $ \\
$ (2, 3,3,10) $  &  $104$ &   $ > 5.08{\cdot} 10^8 $    &     $ \mathbf{368} $  / $ \mathbf{0.00} $  &     $ 370 $  / $ 0.03 $  &     $ 372 $  / $ 7.06 $ \\
$ (2, 3,3,11) $  &  $1499$ &   $ > 5.56{\cdot} 10^8 $    &     $ \mathbf{6824} $  / $ \mathbf{0.23} $  &     $ 6905 $  / $ 0.24 $  &     $ 6892 $  / $ 26.29 $ \\
$ (2, 3,4,16) $  &  $150$ &  $ > 0 $    &     $ \mathbf{576} $  / $ 0.72 $  &     $ \mathbf{576} $  / $ \mathbf{0.70} $  & $-$  \\
$ (2, 3,4,17) $  &  $8236$ &  $ > 0 $    &     $ \mathbf{43368} $  / $ \mathbf{12.43} $  &     $ 43512 $  / $ 12.82 $  & $-$  \\
$ (2, 3,5,25) $  &  $27280$ &  $ > 0 $    &     $ \mathbf{1.61{\cdot} 10^5} $  / $ \mathbf{1166.94} $  &     $ \mathbf{1.61{\cdot} 10^5} $  / $ 1178.14 $  & $-$  \\
$ (2, 4,2,5) $  &  $5$ &  $ 1920 $    &     $ \mathbf{10} $  / $ \mathbf{0.00} $  &     $ \mathbf{10} $  / $ \mathbf{0.00} $  &     $ \mathbf{10} $  / $ \mathbf{0.00} $ \\
$ (2, 4,2,7) $  &  $333$ &  $ 1.60{\cdot} 10^7 $    &     $ 2285 $  / $ \mathbf{0.04} $  &     $ 2224 $  / $ 0.07 $  &     $ \mathbf{1850} $  / $ \mathbf{0.04} $ \\
$ (2, 4,3,9) $  &  $5$ &  $ 2.61{\cdot} 10^7 $    &     $ 36 $  / $ 0.02 $  &     $ 36 $  / $ \mathbf{0.01} $  &     $ \mathbf{26} $  / $ 1102.30 $ \\
\hline 
\end{tabular}}
\end{center}
\end{table}

\emph{Constrained problems.}
Our second set of experiments was on three benchmark domains:
Equidistant Frequency Permutation Array (EFPA),
Balanced Incomplete Block Designs and 
Covering Array (CA) problems. 
We used the non-Boolean model of EFPA~\cite{hmmncp09} (Table~\ref{t:t1}),
the Boolean matrix model of BIBD~\cite{ffhkmpwcp2002} (Table~\ref{t:t3}) and 
a simple model of CA~\cite{Hnich06} (Table~\ref{t:t4}).
We consider the satisfaction version of the CA problem with
a given number of vectors $b$. 
In all problems instances the $\DLex$, $\SnakeLex_R$ and $\SnakeLex_C$ constraints 
show their effectiveness, leaving only a small fraction of symmetric solutions.
Note that $\SnakeLex_C$ often leaves fewer symmetric solutions.
However, it is significantly slower compared to $\DLex$ and $\SnakeLex_R$
because it tends to prune later (thereby exploring larger search trees). For example, the 
number of failures 
for the $(5,3,3,4)$ EFPA problem is $21766$, $14072$ and  $1129085$ for 
$\DLex$,  $\SnakeLex_R$ and   $\SnakeLex_C$ respectively.
On EFPA problems,  $\SnakeLex_R$ is about twice as fast as $\DLex$ 
and leaves less solutions. On the CA problems
$\DLex$ and $\SnakeLex_R$ show similar results,
while $\DLex$ performs better on BIBD problems in terms
of the number of solution left.

Overall, our results show that $\DLex$ and $\SnakeLex$
prune most of the symmetric solutions.
$\SnakeLex_C$ slightly outperforms $\DLex$ and $\SnakeLex_R$ in terms of the
number of solutions left, but it explores larger search trees and is
about two orders of magnitude slower. However, there is little
difference overall in the amount of symmetry eliminated
by the three methods. 

 
\myOmit{
Finally, we can compare the ratio of the number of solutions 
found by
$\DLex$ and $\SnakeLex$ 
  compared to the number of symmetry classes. 
To identify whether a solution satisfies
the `snake' $\RowColSym$ constraint ~\cite{snakelex} we made a minor modification 
of Theorem~\ref{tm:fpt}. Given a row permutation we sort columns in such a way to obtain the lexicographically smallest 
solution  with respect to the snake $\RowColSym$ constraint. In our experiments,
The $\SnakeLex$ constraint was much slower
than the $\DLex$ constraint in finding all solutions and solved to completion about half of the instances.
However, the $\SnakeLex$ constraint shows a better ratio between symmetric and non symmetric solutions.
Consider, for instance, the $(4, 4,4,5)$ benchmark. The $\SnakeLex$
constraint finds 
approximately $3$ solutions on average in each symmetry class.
By comparison, on the same instance, the $\DLex$ constraint finds more 
than $6$ solutions on average in each symmetry class. 
On all benchmarks that $\SnakeLex$ solves within the time-limit,
it leaves the same or fewer solutions as the  $\DLex$ model.
}

\vspace*{-6pt}
\section{Other related work}

Lubiw proved that any matrix has 
a row and column permutation in
which rows and columns are lexicographically
ordered and gave a nearly linear
time algorithm to compute such a matrix \cite{lubiw}. 
Shlyakhter and Flener {\it et al.} independently
proposed 
eliminating row and column
symmetry  using \DLex 
\cite{ilya01,lex2001,ffhkmpwcp2002}. 
To break some of the remaining symmetry, 
Frisch, Jefferson and Miguel
suggested ensuring that the first row is
less than or equal to all
permutations of all other rows \cite{fjmcp2003}. 
As an alternative to ordering both
rows and columns lexicographically,
Frisch {\it et al.} proposed
ordering the rows lexicographically
but the columns with a multiset
ordering 
\cite{fhkmwijcai2003}.
More recently, Grayland {\it et al.}
have proposed \snakelex , an alternative
to \DLex based on linearizing the
matrix 
in a snake-like way \cite{snakelex}.
An alternative way to break
the symmetry of interchangeable
values is to convert it into a variable
symmetry by channelling into a dual 0/1 viewpoint 
in which $Y_{ij}=1$ iff $X_i=j$, 
and using lexicographical ordering constraints
on the columns of the
0/1 matrix \cite{ffhkmpwcp2002}. 
However, this 
hinders propagation
\cite{wecai2006}. 
Finally, dynamic methods like SBDS have been proposed
to remove symmetry from the search
tree \cite{sbds}. 
Unfortunately, dynamic techniques 
tend not to work well with row and columns
symmetries as the number of symmetries is 
usually too large. 

\section{Conclusions}

We have provided a number of
positive and negative results on dealing
with row and column symmetry. 
To eliminate some
(but not all) symmetry 
we can post static constraints like \DLex
and \snakelex .
On the positive side, we proposed the first polynomial
time method to eliminate \emph{all} row and column
symmetry when the number of rows (or columns)
is bounded. 
On the negative side, we argued that 
\DLex and \snakelex can leave a large
number of symmetric
solutions. In addition, we proved that propagating
\DLex completely is NP-hard. 
Finally, we showed that it is not always safe to 
combine Puget's value symmetry
breaking constraints with row and column
symmetry breaking constraints, correcting 
a claim made in the literature. 


\bibliographystyle{splncs}


\end{document}